\documentclass[10pt,twocolumn,letterpaper]{article}

\usepackage{iccv}
\usepackage{times}
\usepackage{epsfig}
\usepackage{graphicx}
\usepackage{amsmath}
\usepackage{amssymb}
\usepackage{multirow}
\usepackage{verbatim}
\usepackage{color}
\usepackage[table]{xcolor}
\usepackage{bbm}
\usepackage{enumitem}
\usepackage{subcaption}
\newcommand{\bx}{\mathbf{x}}
\newcommand{\by}{\mathbf{y}}
\newcommand{\bz}{\mathbf{z}}
\definecolor{blue}{rgb}{1,0,0}

\usepackage[pagebackref=true,breaklinks=true,letterpaper=true,colorlinks,bookmarks=false]{hyperref}

\iccvfinalcopy 

\def\iccvPaperID{5128} 

\ificcvfinal\fi
\begin{document}

\title{Visual Deprojection: Probabilistic Recovery of Collapsed Dimensions}

\author{Guha Balakrishnan\\
MIT\\
{\tt\small balakg@mit.edu}
\and
Adrian V. Dalca\\
MIT and MGH\\
{\tt\small adalca@mit.edu}
\and
Amy Zhao\\
MIT\\
{\tt\small xamyzhao@mit.edu}
\and
John V. Guttag\\
MIT\\
{\tt\small guttag@mit.edu}
\and
Fr\'edo Durand\\
MIT\\
{\tt\small fredo@mit.edu}
\and
William T. Freeman\\
MIT\\
{\tt\small freeman@mit.edu}
}

\maketitle

\begin{abstract}
We introduce visual deprojection: the task of recovering an image or video that has been collapsed along a dimension. Projections arise in various contexts, such as long-exposure photography, where a dynamic scene is collapsed in time to produce a motion-blurred image, and corner cameras, where reflected light from a scene is collapsed along a spatial dimension because of an edge occluder to yield a 1D video. Deprojection is ill-posed-- often there are many plausible solutions for a given input. We first propose a probabilistic model capturing the ambiguity of the task. We then present a variational inference strategy using convolutional neural networks as functional approximators. Sampling from the inference network at test time yields plausible candidates from the distribution of original signals that are consistent with a given input projection. We evaluate the method on several datasets for both spatial and temporal deprojection tasks. We first demonstrate the method can recover human gait videos and face images from spatial projections, and then show that it can recover videos of moving digits from dramatically motion-blurred images obtained via temporal projection. 
\end{abstract}

\begin{figure}[t!]
\begin{center}
\includegraphics[width=\linewidth]{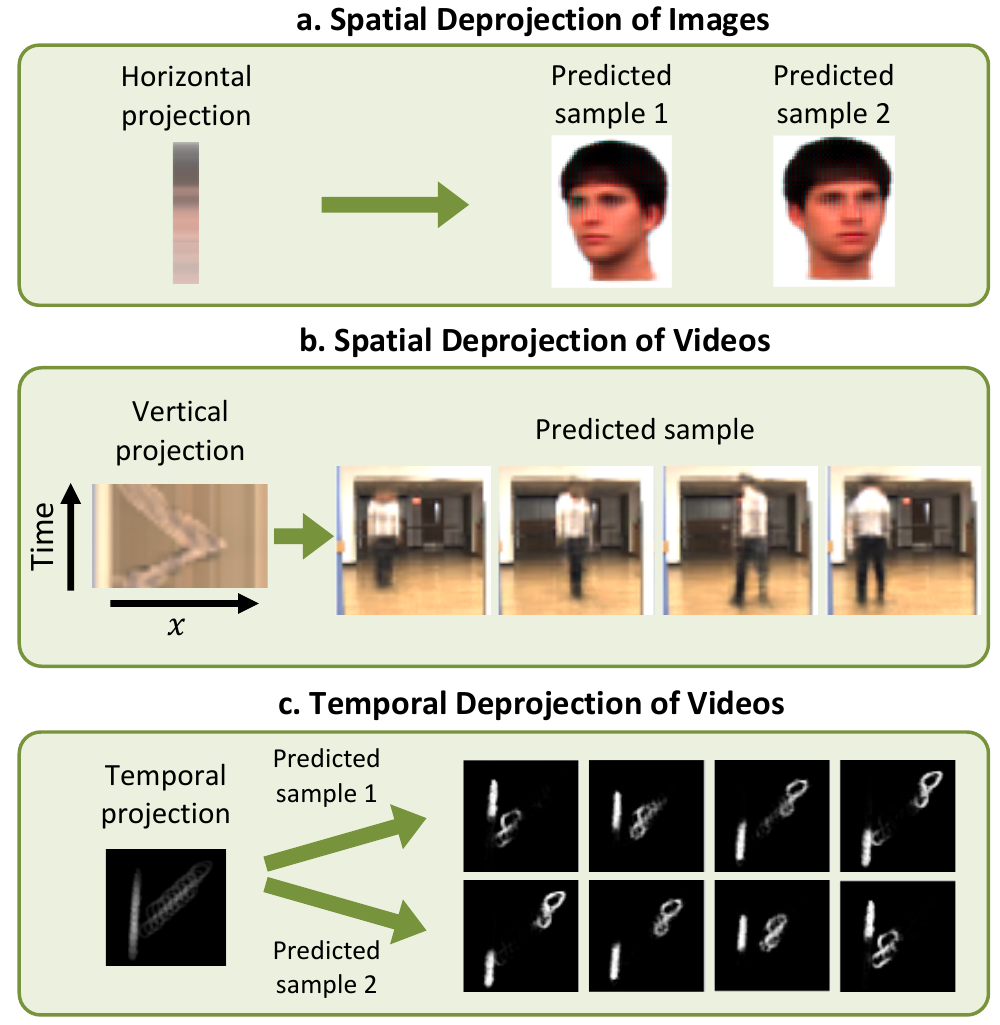}
\end{center}
\caption{Our method takes a spatial projection of an image or video (a, b) or a temporal projection of a video (c), and outputs a distribution over possible original signals. A projection here is an average of pixel values along a dimension of the original signal. The original signal is only one of multiple possible signals that may have plausibly generated that particular projection.}
\label{fig:teaser}
\end{figure}

\section{Introduction}
Captured visual data is often a projection of a higher-dimensional signal ``collapsed'' along some dimension. For example, long-exposure, motion-blurred photographs are produced by projecting motion trajectories along the time dimension~\cite{dai2008motion, Levin_SIGGRAPH2008}. Recent ``corner cameras'' leverage the fact that a corner-like edge occluder vertically projects light rays of hidden scenes to produce a 1D video~\cite{bouman2017turning}. Medical x-ray machines use spatial projectional radiography, where x-rays are distributed by a generator, and the imaged anatomy affects the signal captured by the detector~\cite{logan2016x}. Given projected data, is it possible to synthesize the original signal? In this work, we present an algorithm that enables this synthesis. We focus on recovering images and video from spatial projections, and recovering a video from a long-exposure image obtained via temporal projection. 

The task of inverting projected, high-dimensional signals is ill-posed, making the task infeasible without some priors or constraints on the true signal. This ambiguity includes object orientations and poses in spatial projections, and the ``arrow of time''~\cite{wei2018learning} in temporal projections (Fig.~\ref{fig:teaser}). We leverage the fact that the effective dimension of most natural images is often much lower than the pixel representation, because of the shared structure in a given domain. We handle this ambiguity by formulating a probabilistic model for the generation of signals given a projection. The model consists of parametric functions that we implement with convolutional neural networks (CNNs). Using variational inference, we derive an intuitive objective function. Sampling from this deprojection network at test time produces plausible examples of signals that are consistent with an input projection. 

There is a rich computer vision literature on recovering high-dimensional data from partial observations.  Single-image super-resolution~\cite{freeman2002example}, image demosaicing~\cite{zhen2015image}, and motion blur removal~\cite{fergus2006removing} are all special cases.  Here, we focus on projections where a spatial or temporal dimension is entirely removed, resulting in dramatic loss of information. To the best of our knowledge, ours is the first general recovery method in the presence of a collapsed dimension. We build on insights from related problems to develop a first solution for extrapolating appearance and motion cues (in the case of videos) to unseen dimensions. In particular, we leverage recent advances in neural network-based synthesis and stochastic prediction tasks~\cite{babaeizadeh2017stochastic, isola2017image, xue2018visual}. 

We evaluate our work both quantitatively and qualitatively. We demonstrate that our method can recover the distribution of human gait videos from 2D spacetime images, and face images from their 1D spatial projections.  We also show that our method can model distributions of videos conditioned on motion-blurred images using the Moving MNIST dataset~\cite{srivastava2015unsupervised}. 
\section{Related Work}
\label{sec:rel}
Projections play a central role in computer vision, starting from the initial stages of image formation, where light from the 3D world is projected onto a 2D plane. We focus on a particular class of projections where higher-dimensional signals of interest are collapsed along one dimension to produce observed data.

\subsection{Corner Cameras}
\label{sec:cc}
Corner cameras exploit reflected light from a hidden scene occluded by obstructions with edges to ``see around the corner''~\cite{bouman2017turning}. Reflected light rays from the scene from the same angular position relative to the corner are vertically integrated to produce a 1D video (one spatial dimension + time). That study used the temporal
gradient of the 1D video to coarsely indicate angular positions of the human with respect to the corner, but did not reconstruct the hidden scene. As an initial step towards this difficult reconstruction task, we show that videos and images can be recovered after collapsing one spatial dimension.

\subsection{Compressed Sensing}
Compressed sensing techniques efficiently reconstruct a signal from limited observations by finding solutions to underdetermined linear systems~\cite{candes2008introduction, donoho2006compressed}. This is possible because of the redundancy of natural signals in an appropriate basis. Several methods show that it is possible to accurately reconstruct a signal from a small number ($1000$s) of bases through convex optimization, even when the bases are chosen randomly~\cite{candes2005signal, candes2006near, haupt2006signal}. We tackle an extreme variant where one dimension of a signal is completely lost. We also take a learning-based approach to the problem that yields a distribution of potential signals instead of one estimate.

\subsection{Conditional Image/Video Synthesis and Future Frame Prediction}
Neural network-based image and video synthesis has received significant attention. In conditional image synthesis, an image is synthesized conditioned on some other information, such as a class label or another image of the same dimension (image-to-image translation)~\cite{bousmalis2017unsupervised, isola2017image, odena2016conditional, taigman2016unsupervised, wang2017high, zhu2017unpaired}. In contrast to our work, most of these studies condition on data of the same dimensionality as the output. 

Video synthesis algorithms mainly focus on unconditional generation~\cite{saito2017temporal, tulyakov2017mocogan, vondrick2016generating} or video-to-video translation~\cite{chen2017coherent, shechtman2005space, wang2018vid2vid}. In future video frame prediction, frames are synthesized conditioned on one or more past images. Several of these algorithms treat video generation as a stochastic problem~\cite{babaeizadeh2017stochastic, lee2018stochastic, xue2018visual}, using a variational autoencoder (VAE) style framework~\cite{kingma2013auto}. The inputs and outputs in these problems take a similar form to ours, but the information in the input is different. We draw insights from the stochastic formulation in these studies for our task. 

\subsection{Inverting a Motion-blurred Image to Video}
One application we explore is the formation of videos from dramatically motion-blurred images, created by temporally aggregating photons from a scene over an extended period of time. Two recent studies present the deterministic recovery of a video sequence from a single motion-blurred image~\cite{jin2018learning, purohit2019bringing}. We propose a general deprojection framework for dimensions including, but not limited to time. In addition, our framework is probabilistic, capturing the distribution of signal variability instead of a single deterministic output (see Fig.~\ref{fig:teaser}). 

\section{Methods}
\label{sec:methods}
We assume a dataset of pairs $\{\bx, \by\}$ of original signals \mbox{$\by \in \mathbb{R}^{d_1 \times \cdots \times d_D}$} and projections \mbox{$\bx \in \mathbb{R}^{d_1 \times \cdots d_{p-1} \times d_{p+1} \cdots \times d_D}$}, where $D$ is the number of dimensions of $\by$ and $p$ is the projected dimension. We assume a projection function \mbox{$f_{\omega}: \mathbb{R}^D \rightarrow \mathbb{R}^{D-1}$} with parameters $\omega$. In our experiments, we focus on a case often observed in practice, where $f_{\omega}$ is a linear operation in $\omega$ along $p$, such as averaging: $\bx = f_\omega(\by) = \sum_{k=1}^{d_p} \omega_k \by^{p = k}$, where $\by^{p = k}$ is the $k^{th}$ slice of $\by$ along dimension $p$. For example, a grayscale video \mbox{$\by \in \mathbb{R}^{H\times W \times T}$} might get projected to an image \mbox{$\bx \in \mathbb{R}^{H \times W}$} by averaging pixels across time. Deprojection is a highly underconstrained problem. Even if the values of $\omega$ are known, there are $d_p$ as many variables (size of $\by$) as constraints (size of $\bx$). 

\begin{figure}[t!]
\begin{center}
\includegraphics[width=\linewidth]{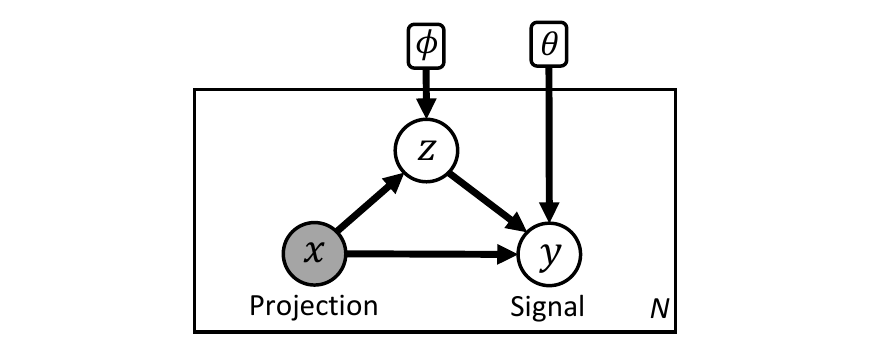}
\end{center}
\caption{Schematic of the probabilistic model at test time. The shaded variable $\bx$ is the observed input projection, $\by$ is the higher-dimensional signal, $\bz$ is a multinomial latent variable, $\phi$ and $\theta$ are global network parameters, and $N$ is the number of test examples in the dataset.}
\label{fig:model}
\end{figure}

We aim to capture the distribution $p(\by | \bx)$ for a particular scenario with data. We first present a probabilistic model for the deprojection task which builds on the conditional VAE (CVAE)~\cite{sohn2015learning} (Fig.~\ref{fig:model}). We let $\bz \sim p_{\phi}(\bz | \bx)$ be a multivariate normal latent variable which captures variability of $\by$ unexplainable from $\bx$ alone. Intuitively, $\bz$ encodes information orthogonal to the unprojected dimensions. For example, it could capture the temporal variation of the various scenes that may have led to a long-exposure image.
  
We define $p_{\theta}(\by | \bx, \bz)$ as a Gaussian distribution:
\begin{align}
p_{\theta}(\by | \bx, \bz) = & \mathcal{N}(\by; g_{\theta}(\bx, \bz), \mathbf{I} \sigma_y^2)
\label{eq:py}
\end{align}
\noindent where $\sigma_y^2$ is a per-pixel noise variance and $g_{\theta}(\bx, \bz)$ is a \emph{deprojection} function, parameterized by $\theta$ and responsible for producing a noiseless estimate of $\by$ given $\bx$ and $\bz$. 

\begin{figure*}[t!]
\begin{center}
\includegraphics[width=\textwidth]{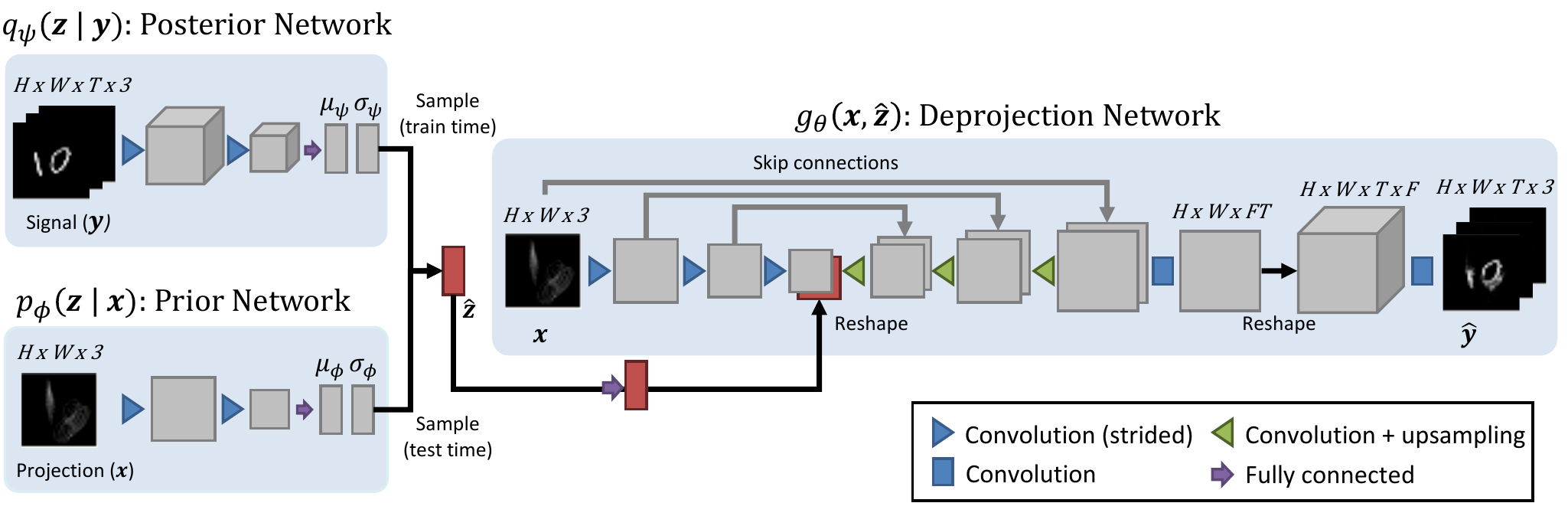}
\end{center}
\caption{Overview of our network architecture, drawn here for the 2D-to-3D temporal deprojection task. The network consists of three parameterized functions: $q_{\psi} (\cdot | \cdot)$ captures the variational posterior distribution, $p_{\phi} (\cdot | \cdot)$ captures the prior distribution and $g_{\theta} (\cdot, \cdot)$ performs deprojection. $\mathbf{z}$ is sampled from $q_{\psi} (\cdot)$ during training, and from $p_{\phi}(\cdot)$ during testing.}
\label{fig:network}
\end{figure*}

\subsection{Variational Inference and Loss Function}
Our goal is to estimate $p_{\phi, \theta}(\mathbf{y}|\mathbf{x})$:
\begin{flalign}
p_{\phi, \theta}(\mathbf{y}|\mathbf{x}) &= \int_{\bz} p_{\theta}(\mathbf{y} | \bx, \bz)p_{\phi}(\bz | \bx)d\bz
\end{flalign}
\noindent Evaluating this integral directly is intractable because of its reliance on potentially complex parametric functions and the intractability of estimating the posterior $p(\bz | \by)$. We instead use variational inference to obtain a lower bound of the likelihood, and use stochastic gradient descent to optimize it~\cite{jordan1999introduction, kingma2013auto}. We introduce an approximative posterior distribution $q_{\psi}(\mathbf{z}|\mathbf{y}) = \mathcal{N}(\bz; \mu_\psi (\by), \sigma_\psi (\by))$:
\begin{flalign}
\log p_{\phi, \theta}(\mathbf{y}|\mathbf{x}) &= \log E_{\mathbf{z} \sim q_{\psi}}\left[ \frac{p_{\phi}(\bz | \bx)}{q_{\psi}(\mathbf{z}|\mathbf{y})} p_{\theta}(\mathbf{y} | \mathbf{x}, \mathbf{z})\right].
\end{flalign}
Using Jensen's inequality, we achieve the following evidence lower bound (ELBO) for $\log p_{\phi, \theta}(\mathbf{y}|\mathbf{x})$:
\begin{flalign}
\log p_{\phi, \theta}(\mathbf{y}|\mathbf{x}) \geq E_{\mathbf{z} \sim q_{\psi}} \big[\log p_{\theta}(\mathbf{y} | \mathbf{x}, \mathbf{z}) \big] \\
- D_{KL}[q_{\psi}(\mathbf{z}|\mathbf{y}) || p_{\phi}(\mathbf{z} | \mathbf{x})],  \nonumber
\label{eq:elbo}
\end{flalign}
\noindent where $D_{KL}[\cdot || \cdot ]$ is the Kullback-Leibler divergence encouraging the variational distribution to approximate the conditional prior, resulting in a regularized embedding. We estimate the expectation term by drawing one $\hat{\mathbf{z}}$ from $q_{\psi}(\mathbf{z}|\mathbf{y})$ within the network using the reparametrization trick~\cite{kingma2013auto} and evaluating the expression: 
\begin{flalign}
\log p_{\theta}(\mathbf{y} | \mathbf{x}, \hat{\mathbf{z}}) =& \frac{||g_{\theta}(\mathbf{x}, \hat{\bz}) - \mathbf{y}||_2^2}{2\sigma_\mathbf{y}^2} + \text{const}.
\end{flalign}
This leads to the training loss function to be minimized:
\begin{flalign}
\mathcal{L}_{\phi, \psi, \theta} (\mathbf{x}, \mathbf{y}, \hat{\bz})=&\beta D_{KL}[q_{\psi}(\bz|\by) || {p_{\phi}(\bz | \bx)}] \nonumber \\
& + {||g_{\theta}(\mathbf{x}, \hat{\bz}) - \by||^2_2}
\label{eq:loss}
\end{flalign}
\noindent where $\beta$ is a tradeoff parameter capturing the relative importance of the regularization term. The per-pixel reconstruction term in Eq.~\eqref{eq:loss} can result in blurry outputs. For datasets with subtle details such as face images, we also add a perceptual error, computed over a learned feature space~\cite{dosovitskiy2016generating, johnson2016perceptual, zhang2018unreasonable}. We use a distance function $D_{\gamma}(\cdot, \cdot)$~\cite{zhang2018unreasonable}, computed over high-dimensional features learned by the VGG16 network~\cite{simonyan2014very} with parameters $\gamma$, trained to perform classification on ImageNet. 

\subsection{Network Architectures}
We implement $g_{\theta} (\cdot, \cdot)$ and the Gaussian parameters of $q_{\psi}(\cdot | \cdot)$ and $p_{\phi}(\cdot | \cdot)$ with neural networks. Fig.~\ref{fig:network} depicts the architecture for the 2D-to-3D temporal deprojection task. Our 2D-to-3D spatial deprojection architecture is nearly identical, differing only in the dimensions of $\bx$ and the reshaping operator's dimension ordering. We handle 1D-to-2D deprojections by using the lower-dimensional versions of the convolution and reshaping operators. The number of convolutional layers, and number of parameters vary by dataset based on their complexities.

\subsubsection{Posterior and Prior Encoders}
The encoder for the distribution parameters of the posterior $q_{\psi}(\cdot | \cdot)$ is implemented using a series of strided 3D convolutional operators and Leaky ReLU activations until a volume of resolution less than $8\times 8 \times 3$ is reached. We flatten this volume and use two fully connected layers to obtain $\mu_{\psi}$ and $\sigma_{\psi}$, the distribution parameters. The encoder for the conditional prior $p_{\phi}(\cdot | \cdot)$ is implemented in a similar way, with 2D strided convolutions. One $\hat{\bz}$ is drawn from $q_{\psi}(\cdot | \cdot)$ and fed to the deprojection function. At test time, $\hat{\bz}$ is drawn from $p_{\phi} (\cdot | \cdot)$ to visualize results. 

\subsubsection{Deprojection Function}
The function $g_\theta(\bx,\hat{\mathbf{z}})$ deprojects $\bx$ into an estimate $\hat{\by}$. We first use a UNet-style architecture~\cite{ronneberger2015u} to compute per-pixel features of $\bx$. The UNet consists of two stages. In the first stage, we apply a series of strided 2D convolutional operators to extract multiscale features. We apply a fully connected layer to $\hat{\bz}$, reshape these activations into an image, and concatenate this image to the coarsest features. The second stage applies a series of 2D convolutions and upsampling operations to synthesize an image of the same dimensions as $\bx$ and many more data channels. Activations from the first stage are concatenated to the second stage activations with skip connections to propagate learned features.

We expand the resulting image along the collapsed dimension to produce a 3D volume. To do this, we apply a 2D convolution to produce $TF$ data channels, where $T$ is the size of the collapsed dimension (time in this case), and $F$ is some number of features. Finally, we reshape this image into a 3D volume, and apply a few 3D convolutions to refine and produce a signal estimate $\hat{\by}$. 
\section{Experiments and Results}
\label{sec:experiments}
We first evaluate our method on 1D-to-2D spatial deprojections of human faces using FacePlace~\cite{righi2012recognizing}. We then show results for 2D-to-3D spatial deprojections using an in-house dataset of human gait videos collected by the authors. Finally, we demonstrate 2D-to-3D temporal deprojections using the Moving MNIST~\cite{srivastava2015unsupervised} dataset. We focus on projections where pixels are averaged along a dimension for all experiments. For all experiments we split the data into train/test/validation non-overlapping groups. 

\subsection{Implementation}
We implement our models in Keras~\cite{chollet2015keras} with a Tensorflow~\cite{abadi2016tensorflow} backend. We use the ADAM optimizer~\cite{kingma2014adam} with a learning rate of $1e^{-4}$. We trained separate models for each experiment. We select the regularization hyperparameter $\beta$ separately for each dataset such that the KL term is between $[5,15]$ on our validation data, to obtain adequate data reconstruction while avoiding mode collapse. We set the dimension of $\bz$ to 10 for all experiments. 

\begin{figure}[t!]
\begin{subfigure}{\linewidth}
\includegraphics[width=\linewidth]{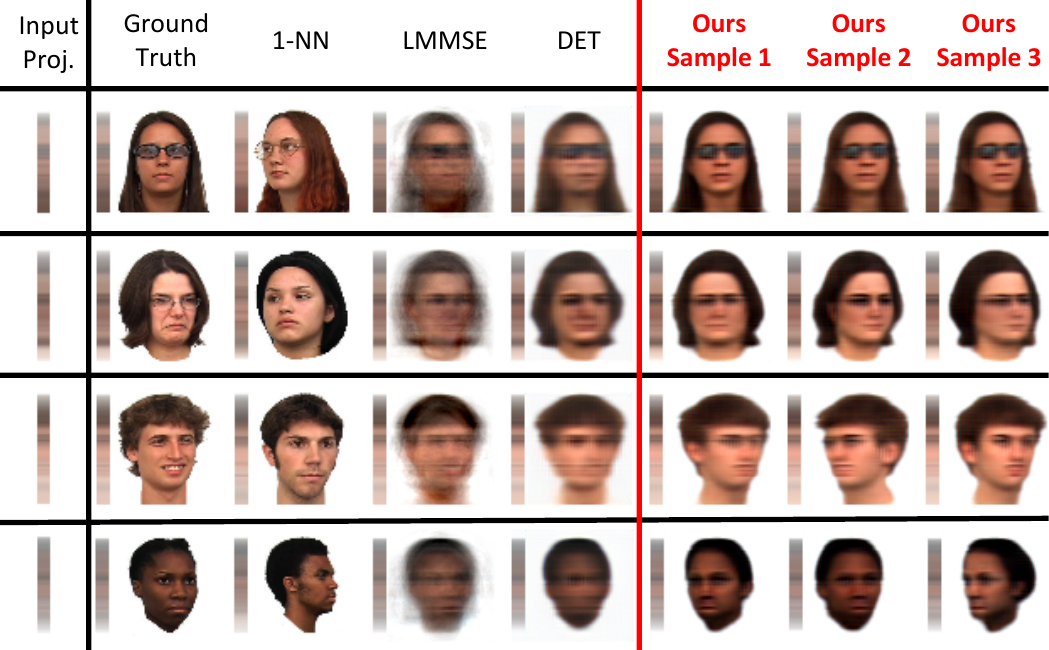}
\caption{Horizontal projection results.} \label{fig:faces_horizontal}
\end{subfigure}
\begin{subfigure}{\linewidth}
\includegraphics[width=\linewidth]{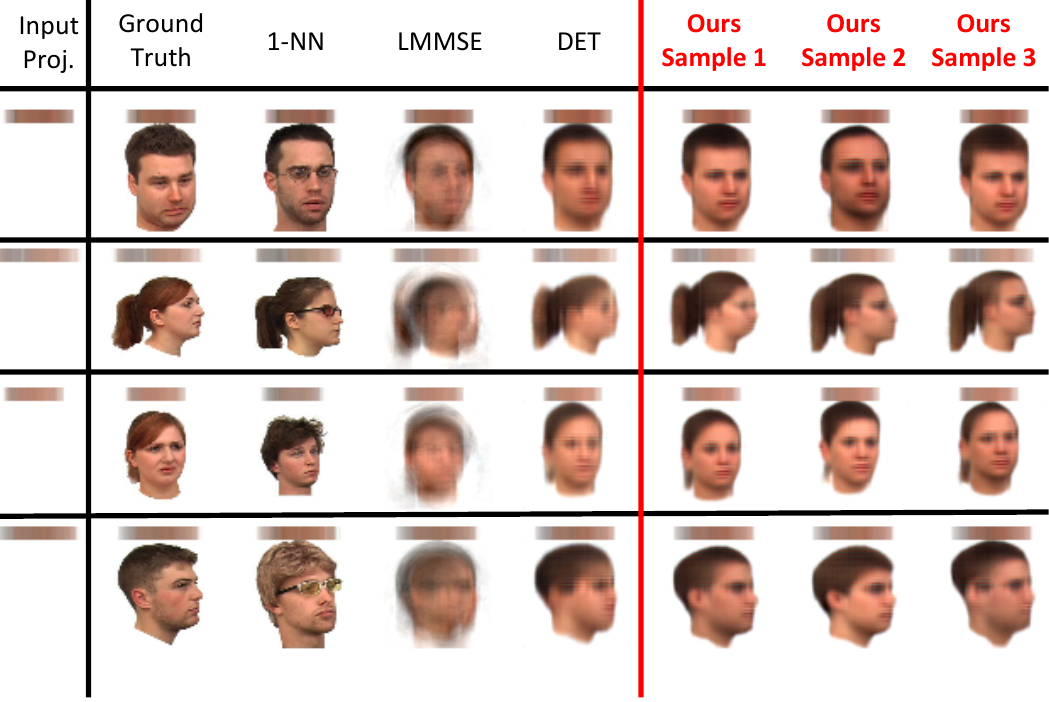}
\caption{Vertical projection results.} \label{fig:faces_vertical}
\end{subfigure}
\caption{Sample image reconstructions on FacePlace. The input projections along with ground truth images are shown on the left. Our method's samples are randomly chosen. Our method is able to synthesize a variety of appearances with projections closely matching the input.} \label{fig:faces}
\end{figure}

\begin{figure}[t!]
	\centering
	\begin{minipage}[t]{8cm}
		\centering
		\includegraphics[width=\linewidth]{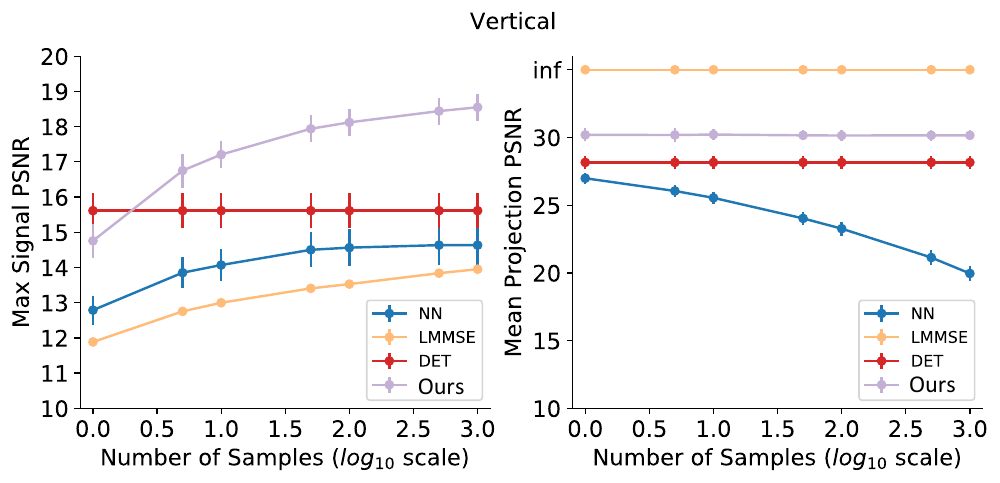}
	\end{minipage}
	\begin{minipage}[t]{8cm}
		\centering
		\includegraphics[width=\linewidth]{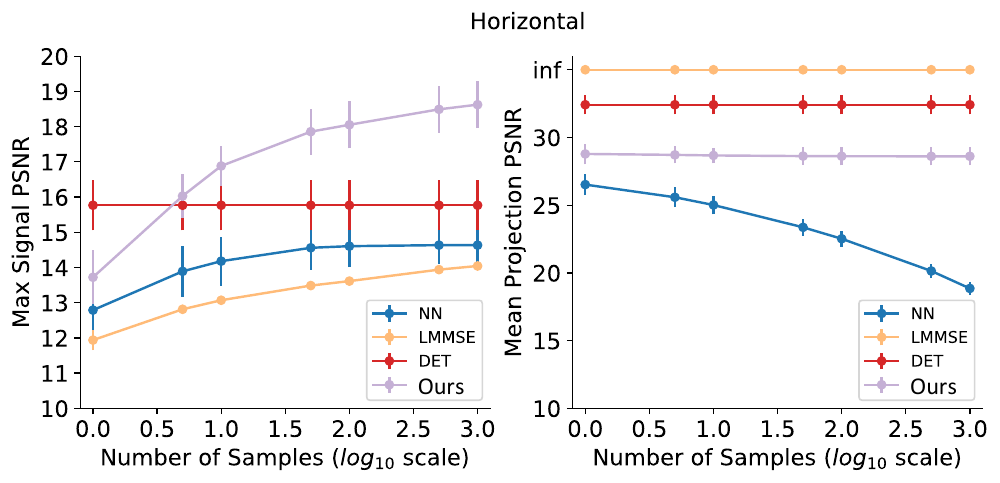}
	\end{minipage}
	\caption{FacePlace PSNR for all methods (vertical projection on top, horizontal on bottom, max signal PSNR (deprojection estimate) on left, mean projection PSNR on right) with varying sample size for 100 test projections. Our method yields higher maximum signal PSNR than all baselines. DET has a higher expected signal PSNR for one sample because it tends to return a blurry average over many signals. LMMSE has infinite projection PSNR because it captures the exact linear signal-projection relationship by construction.}
\label{fig:face_psnr}
\end{figure}
\begin{figure*}[t!]
	\centering
	\includegraphics[width=\linewidth]{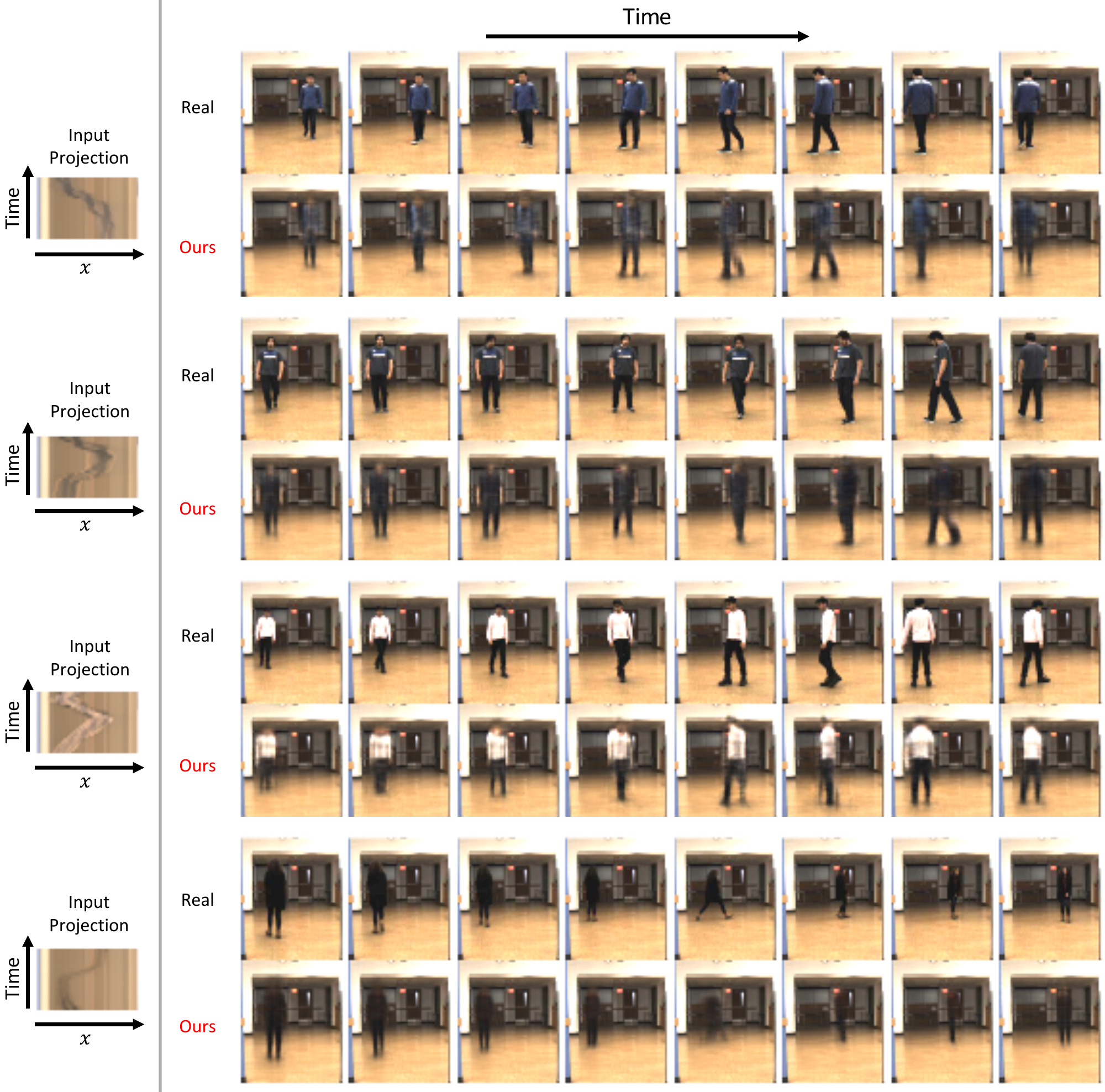}
	\caption{Sample outputs for four examples from the in-house walking dataset. The left column shows the input vertical projection. For each example, the top row displays the ground truth sequence and the bottom row displays our method's mean output using $\bz = \mu_{\phi}$.}
\label{fig:walk_synth_results}
\end{figure*}
\subsection{Spatial Deprojections with FacePlace}
FacePlace consists of over 5,000 images of 236 different people. There are many sources of variability, including different ethnicities, multiple views, facial expressions, and props. We randomly held out all images for 30 individuals to form a test set. We scaled images to $128\times128$ pixels and performed data augmentation with translation, scaling and saturation variations. We compare our method against the following baselines:

\begin{enumerate}[itemsep=0ex, topsep=0pt]
\item Nearest neighbor selector ($k$-NN): Selects the $k$ images from the training dataset with projections closest to the test projection using mean squared error distance. 
\item A deterministic model (DET) identical to the deprojection network $g_{\theta}(\bx,\bz)$ of our method, without the incorporation of a latent variable $\bz$.
\item A linear minimum mean squared error (LMMSE) estimator which assumes that $\bx$ and $\by$ are drawn from distributions $X, Y$ such that $\bar{\by} = E_Y[\by]$ is linear in $\bx$: $\bar{\by} = A\bx + b$ for some parameters $A$ and $b$. Minimizing the expected MSE of $\by$ yields a closed form expression for $p(\by | \bx)$:
\begin{align}
p(\by | \bx)= \mathcal{N}(\by; &\Sigma_{YX}\Sigma_X^{-1}(\bx - \bar{\bx}) + \bar{\by}, \nonumber \\
&\Sigma_{Y} - \Sigma_{YX}\Sigma_X^{-1}\Sigma^T_{YX}), 
\label{eq:mmse}
 \end{align}
where $\Sigma_X$ and $\Sigma_Y$ are the covariance matrices of $X$ and $Y$ and $\Sigma_{XY}$ is their cross-covariance matrix.\\
\end{enumerate}

For both our method and DET, we used the perceptual loss metric. Fig.~\ref{fig:faces} presents visual results, with a few randomly chosen samples from our method. $1$-NN varies in performance depending on the test example, and can sometimes produce faces from the wrong person. LMMSE produces very blurry outputs, indicating the highly nonlinear nature of this task. DET produces less blurry outputs, but still often merges different plausible faces together. Our method captures uncertainty of head orientations as well as appearance variations, such as hair color and facial structure. Ambiguity in head orientation is more apparent with the horizontal projections, since pose changes affect that dimension the most. The outputs of our methods are also sharper than LMMSE and DET, and are more consistent with ground truth than $1$-NN.

We also quantitatively evaluate the models. We use PSNR (peak-signal-to-noise-ratio, higher is better) to measure reconstruction quality between images. For each test projection, we sample $k$ deprojection estimates from each model (DET always returns the same estimate) and record the highest PSNR between any estimate and the ground truth image. For each deprojection estimate, we reproject and record the average PSNR of the output projections with respect to the the test (initial) projection. 

Fig.~\ref{fig:face_psnr} illustrates the results with varying samples $k$ for 100 test projections. As the number of samples $k$ increases, our method's signal (deprojection) PSNR improves, highlighting the advantage of our probabilistic approach. Best estimates from $k$-NN approach the best estimates of our method in signal reconstruction with increasing $k$, but many poor estimates are also retrieved by $k$-NN as evidenced by its decreasing projection PSNR curve. LMMSE has perfect projection PSNR because it captures the exact linear relationship between the signal and projection by construction. DET has higher signal PSNR when drawing one sample, because it averages over plausible images, while our method does not. Our proposed method surpasses DET after 1 sample.

\subsection{Spatial Deprojections with Walking Videos}
We qualitatively evaluate our method on reconstructing human gait videos from vertical spatial projections. This scenario is of practical relevance for corner cameras, described in Sec.~\ref{sec:cc}. We collected 35 videos of 30 subjects walking in a specified area for one minute each. Subjects had varying attire, heights (5'2''- 6'5''), ages (18-60), and sexes (18m/12f). Subjects were not instructed to walk in any particular way, and many walked in odd patterns. The background is identical for all videos. We downsampled the videos to 5 frames per second and each frame to $256\times 224$ pixels, and apply data augmentation of horizontal translations to each video. We held out 6 subjects to produce a test set. We predict sequences of 24 frames (roughly 5 seconds in real time).

Fig.~\ref{fig:walk_synth_results} presents several reconstruction examples, obtained by setting $\hat{\bz} = \mu_\phi (\bx)$, the mean of the prior distribution. Our method recovers many details from the vertical projections alone. The background is easily synthesized because it is consistent among all videos in the dataset. Remarkably, many appearance and pose details of the subjects are also recovered. Subtle fluctuations in pixel intensity and the shape of the projected foreground trace contain clues about the foreground signal along the collapsed dimension. For example, the method seems to learn that a trace that gets darker and wider with time likely corresponds to a person walking closer to the camera.

The third subject is an illustrative result for which our method separates the white shirt from black pants despite their aspects not being obvious in the projection. Projected details, along with a learned pattern that shirts are often lighter colors than pants, likely enable this recovery. Finally, the method may struggle with patterns rarely seen in the training data, such as the large step by the fourth subject in the fifth frame.

\begin{figure}[b!]
	\centering
	\includegraphics[width=\linewidth]{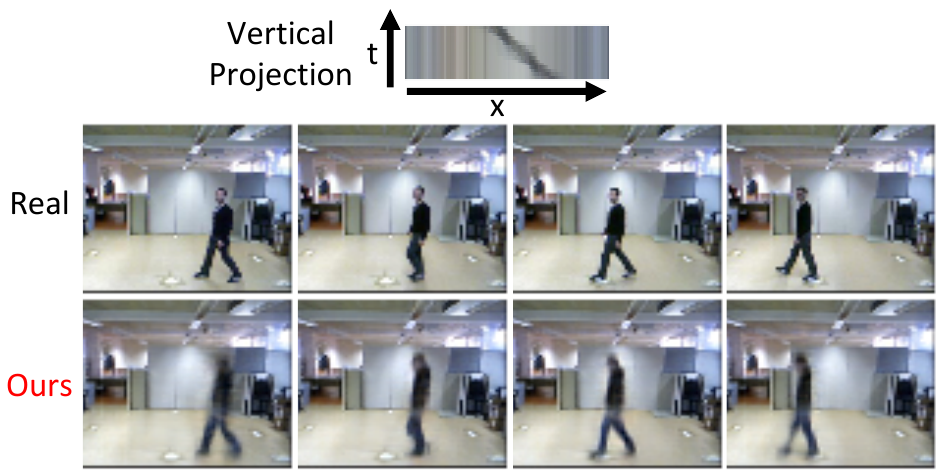}
	\caption{Sample output from the DGAIT walking dataset.}
\label{fig:gait_results}
\end{figure}

In addition to these experiments, we trained a separate model on the DGAIT dataset~\cite{borras2012depth} consisting of more subjects (53), but with simpler walking patterns. We obtain results with similar quality, as illustrated in Fig.~\ref{fig:gait_results}.

\subsection{Temporal Deprojections with Moving MNIST}
The Moving MNIST dataset consists of $10,000$ video sequences of two moving handwritten digits. The digits can occlude one another, and bounce off the edges of the frames. Given a dataset of $64\times64\times10$-sized video subclips, we generate each projection $\bx$ by averaging the frames in time, similar to other studies that generate motion-blurred images at a large scale~\cite{jin2018learning, kim2017online, nah2017deep, noroozi2017motion}. Despite the simple appearance and dynamics of this dataset, synthesizing digit appearances and capturing the plausible directions of each trajectory is challenging. 

\begin{figure}[t!]
	\centering
	\includegraphics[width=\linewidth]{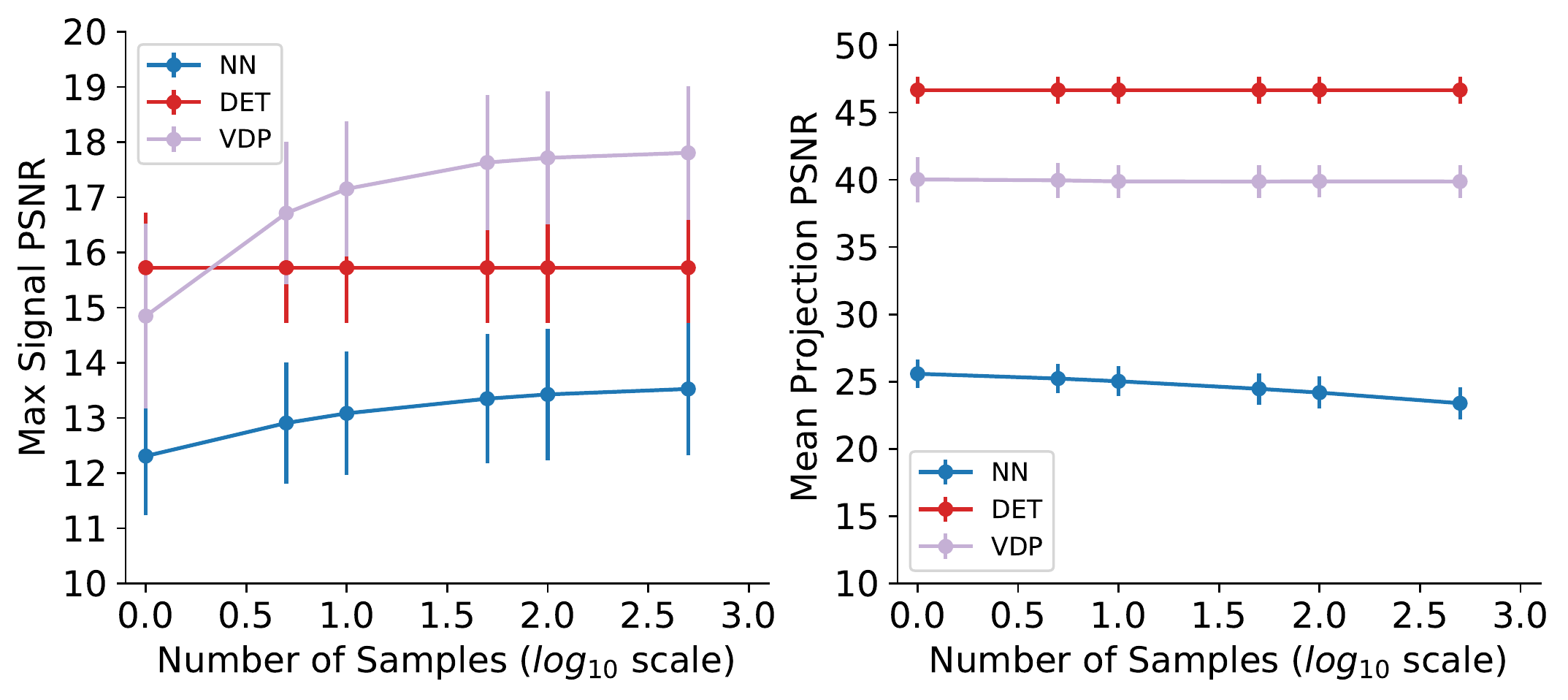}
	\caption{Moving MNIST PSNR plots for 100 projection test examples, similar to Fig.~\ref{fig:face_psnr}.}
\label{fig:moving_mnist_psnr}
\end{figure}

\begin{figure*}[t!]
	\centering
	\includegraphics[width=\linewidth]{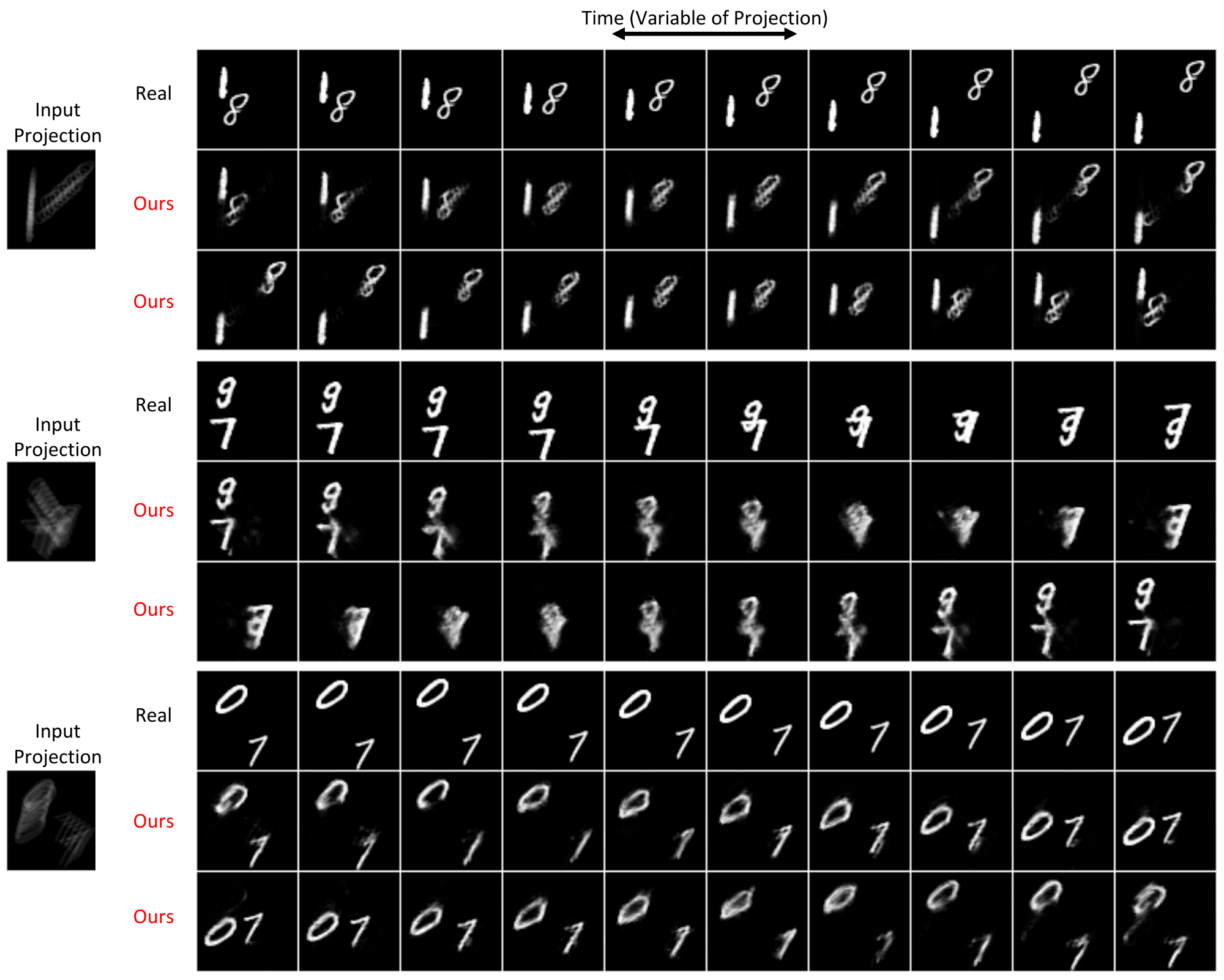}
	\caption{Sample outputs from the Moving MNIST dataset. The left column shows the input projection. For each example, the top row displays the ground truth sequence. We show two sample sequences produced by our method per input projection: the first matches the temporal direction of the ground truth, and the second synthesizes the reverse temporal progression.}
\label{fig:moving_mnist}
\end{figure*}

Sample outputs of our method for three test examples are visualized in Fig.~\ref{fig:moving_mnist}. To illustrate the temporal aspects learned by our method, we sample 10 sequences from our method for each projection, and present the sequences with the lowest mean squared error with respect to the ground truth clip run forwards and backwards. Our method is able to infer the shape of the characters from a dramatically motion-blurred input image, difficult to interpret even by human standards. Furthermore, our method captures the multimodal dynamics of the dataset, which we illustrate by presenting the two motion sequences: the first sequence matches the temporal direction of the ground truth, and the second matches the reverse temporal progression. 

We quantify our accuracy using PSNR curves, similar to the first experiment, displayed in Fig.~\ref{fig:moving_mnist_psnr}. Because of the prohibitive computational costs of generating the full joint covariance matrix, we do not evaluate LMMSE in this experiment. DET produces blurry sequences, by merging different plausible temporal orderings. Similar to the first experiment, this results in DET outputs having the best expected signal (deprojection) PSNR only for $k=1$. Our method clearly outperforms DET in signal PSNR for $k > 1$. DET performs better in projection PSNR, since in this experiment an average of all plausible sequences yields a very accurate projection. $k$-NN performs relatively worse in this experiment compared to the FacePlace experiments, because of the difficulty in finding nearest neighbors in higher-dimensions.  
\section{Conclusion}
\label{sec:conclusion}
In this work, we introduced the novel problem of \textit{visual deprojection}: synthesizing an image or video that has been collapsed along a dimension into a lower-dimensional observation. We presented a first general method that handles both images and video, and projections along any dimension of these data. We addressed the uncertainty of the task by first introducing a probabilistic model that captures the distribution of original signals conditioned on a projection. We implemented the parameterized functions of this model with CNNs to learn shared image structures in each domain and enable accurate signal synthesis. 

Though information from a collapsed dimension is often seemingly unrecoverable from a projection to the naked eye, our results demonstrate that much of the ``lost'' information is recoverable. We demonstrated this by reconstructing subtle details of faces in images and accurate motion in videos from spatial projections alone. Finally, we illustrate that videos can be reconstructed from dramatically motion blurred images, even with multimodal trajectories, using the Moving MNIST dataset. This work illustrates promising results in a new, ambitious imaging task and opens exciting possibilities in future applications of revealing the invisible. 

\subsubsection*{Acknowledgments}
\noindent This work was funded by DARPA REVEAL Program under Contract No. HR0011-16-C-0030, NIH 1R21AG050122 and Wistron Corp.

{\small
\bibliographystyle{ieee_fullname}
\bibliography{egbib}
}

\end{document}